\documentclass{article}
\usepackage{mathrsfs,amsmath,amsthm,amssymb,bm,algorithmic,algorithm,epsfig}
\newtheorem{thm}{Theorem}
\newtheorem{lem}[thm]{Lemma}
\newtheorem{cor}[thm]{Corollary}

\title{Metric Embedding for Nearest Neighbor Classification}

\author{
Bharath K. Sriperumbudur \& Gert R. G. Lanckriet\\
Department of Electrical and Computer Engineering\\
University of California, San Diego\\
La Jolla, CA 92093. \\
\texttt{\{bharathsv@ucsd.edu, gert@ece.ucsd.edu\}} \\
}
\begin{document}
\maketitle
\begin{abstract}
The distance metric plays an important role in nearest neighbor
(NN) classification. Usually the Euclidean distance metric is
assumed or a Mahalanobis distance metric is optimized to improve
the NN performance. In this paper, we study the problem of
embedding arbitrary metric spaces into a Euclidean space with the
goal to improve the accuracy of the NN classifier. We propose a
solution by appealing to the framework of regularization in a
reproducing kernel Hilbert space and prove a representer-like
theorem for NN classification. The embedding function is then
determined by solving a semidefinite program which has an
interesting connection to the soft-margin linear binary support
vector machine classifier. Although the main focus of this paper
is to present a general, theoretical framework for metric
embedding in a NN setting, we demonstrate the performance of the
proposed method on some benchmark datasets and show that it
performs better than the Mahalanobis metric learning algorithm in
terms of leave-one-out and generalization errors.
\end{abstract}
\section{Introduction}\label{Sec:Introduction}
The nearest neighbor (NN) algorithm~\cite{Fix-51,Cover-67} is one
of the most popular non-parametric supervised classification
methods. Because of the non-linearity of its decision boundary,
the NN algorithm generally provides good classification
performance. The algorithm is straight forward to implement and is
easily extendible to multi-class problems unlike other popular
classification methods like support vector machines
(SVM)~\cite{Vapnik-95}. The $k$-NN rule classifies each unlabelled
example by the majority label among its $k$-nearest neighbors in
the training set. Therefore, the performance of the rule depends
on the distance metric used, which defines the nearest neighbors.
In the absence of prior knowledge, the examples are assumed to lie
in a Euclidean metric space and the Euclidean distance is used to
find the nearest neighbor. However, often there are distance
measures that better reflect the underlying structure of the data
at hand, which, if used, would lead to better NN classification
performance. \par Let $(\mathscr{X},\rho)$ represent a metric
space $\mathscr{X}$ (or more generally, a semimetric space) with
$\rho:\mathscr{X}\times\mathscr{X}\rightarrow \mathbb{R}^+$ as its
metric (semimetric) and $x\in\mathscr{X}$. For example, (a) when
$x$ is an image, $\mathscr{X}=\mathbb{R}^d$ and $\rho$ is the
\emph{tangent distance} between images, (b) for $x$ lying on a
manifold, $\mathscr{X}=\text{manifold in } \mathbb{R}^d$ and
$\rho$ is the \emph{geodesic distance}, and (c) in a structured
setting like a graph, $\mathscr{X}=\{\text{vertices}\}$ and
$\rho(x,y)$ is the \emph{shortest path distance} from $x$ to $y$,
where $x,y\in\mathscr{X}$. These settings are practical with (a)
and (b) more prominent in computer vision and (c) in
bio-informatics. However, in such scenarios, the true underlying
distance metric may not be known or it might be difficult to
estimate it for its use in NN classification. In such cases, as
aforementioned, often the Euclidean distance metric is used
instead. The goal of this paper is to extend the NN rule to
arbitrary metric spaces, $\mathscr{X}$, wherein we propose to
embed the given training data into a space whose underlying metric
is known.
\par Prior works~\cite{Xing-02,Shalev-04,NCA,Kilian-05} deal with
$\mathscr{X}=\mathbb{R}^D$ and assume the Mahalanobis distance
metric, i.e.,
$\rho(\mathbf{x},\mathbf{y})=\sqrt{(\mathbf{x}-\mathbf{y})^T\mathbf{A}(\mathbf{x}-\mathbf{y})}$
(with $\mathbf{A}\succeq 0$), which is then optimized with the
goal to improve NN classification performance. These methods can
be interpreted as finding a linear transformation
$\mathbf{L}\in\mathbb{R}^{d\times D}$ so that the transformed data
lie in a Euclidean metric space, i.e.,
$\rho(\mathbf{x},\mathbf{y})=\sqrt{(\mathbf{x}-\mathbf{y})^T\mathbf{A}(\mathbf{x}-\mathbf{y})}=||\mathbf{Lx}-\mathbf{Ly}||_2$
with $\mathbf{A}=\mathbf{L}^T\mathbf{L}$. All these methods learn
the Mahalanobis distance metric by minimizing the distance between
training data of same class while separating the data from
different classes with a large margin. Instead of assuming the
Mahalanobis distance metric which restricts $\mathscr{X}$ to
$\mathbb{R}^D$, we would like to find some general transformation
(instead of linear) that embeds the training data from an
arbitrary metric space, $\mathscr{X}$ into a Euclidean space while
improving the NN classification performance.
\par In this paper, we propose to minimize the proxy to average leave-one-out error
(LOOE) of a $\varepsilon$-neighborhood NN classifier by embedding
the data into a Euclidean metric space. To achieve this, we study
two different approaches that learn the embedding function. The first approach deals within the framework of regularization in
a reproducing kernel Hilbert space (RKHS)~\cite{Scholkopf-02},
wherein $f\in\mathscr{H}_k=\{f\,|\,
f:\mathscr{X}\rightarrow\mathbb{R}^d\}$ is learned with
$k:\mathscr{X}\times\mathscr{X}\rightarrow\mathbb{R}$ being the
reproducing kernel of $\mathscr{H}_k$. We prove a
\emph{representer-like theorem} for NN classification and show
that $f$ admits the form $f=\sum^n_{i=1}\mathbf{c}_i k(.,x_i)$
with $\sum^n_{i=1}\mathbf{c}_i=\mathbf{0}$, where
$\{\mathbf{c}_i\}^n_{i=1}\in\mathbb{R}^d$ and $n$ is the number of
training points. Therefore, the problem of learning $f$ reduces to
learning $\{\mathbf{c}_i\}^n_{i=1}$, resulting in a non-convex
optimization problem which is then relaxed to yield a convex
semidefinite program (SDP)~\cite{sdp}. We provide an interesting
interpretation of this approach by showing that the obtained SDP
is in fact a soft-margin linear binary SVM classifier. In the
second approach, we learn a Mercer kernel map
$\phi:\mathscr{X}\rightarrow\ell^N_2$ that satisfies $\langle
\phi(x),\phi(y)\rangle=k(x,y),\,\forall x,y\in\mathscr{X}$, where
$k$ is the Mercer kernel and $N\in\mathbb{N}$ or $N=\infty$
depending on the number of non-zero eigenvalues of $k$. We show
that learning $\phi$ is equivalent to learning the kernel $k$.
However, the learned $k$ is not interesting as it does not allow
for an out-of-sample extension and so can be used only in a
transductive setting. Using the algorithm derived from the RKHS
framework, some experiments are carried out on four benchmark
datasets, wherein we show that the proposed method has better
leave-one-out and generalization error performance compared to the
Mahalanobis metric learning algorithm proposed
in~\cite{Kilian-05}.
\section{Problem formulation}\label{Sec:ProblemFormulation}
Let $\{x_i,y_i\}^n_{i=1}$ denote the training set of $n$ labelled
examples with $x_i\in\mathscr{X}$ and $y_i\in\{1,2,\ldots,l\}$,
where $l$ is the number of classes. Unlike prior works which learn
a linear transformation
$\mathbf{L}:\mathbb{R}^D\rightarrow\mathbb{R}^d$ (assuming
$\mathscr{X}=\mathbb{R}^D$), leading to the distance metric,
$\rho_{\mathbf{L}}(x_i,x_j)=||\mathbf{L}x_i-\mathbf{L}x_j||_2$,
our goal is to learn a transformation,
$g\in\mathscr{G}=\{g\,|\,g:\mathscr{X}\rightarrow\mathscr{Y}\}$ so
that (a) the average LOOE of the $\varepsilon$-neighborhood NN
classifier is reduced and (b) $\mathscr{Y}$ is Euclidean,
i.e., $\rho_{g}(x_i,x_j)=||g(x_i)-g(x_j)||_2$.
\par Let $\mathbb{B}_g(x,\varepsilon)=\{g(y):\rho^2_g(x,y)\le\varepsilon\}$ represent a Euclidean ball centered at $g(x)$ with radius
$\sqrt{\varepsilon}$. Let $\tau_{ij}=2\delta_{y_i,y_j}-1$ where
$\delta$ represents the Kronecker delta.\footnote{The Kronecker
delta is defined as $\delta_{ij}=1$, if $i=j$ and $\delta_{ij}=0$,
if
$i\ne j$.}
Let $\mu_x(\mathbb{A})$ denote a Dirac measure\footnote{The Dirac
measure for any measurable set $\mathbb{A}$ is defined by
$\mu_x(\mathbb{A})=1$, if $x\in\mathbb{A}$ and
$\mu_x(\mathbb{A})=0$, if $x\notin\mathbb{A}$.} for any measurable set $\mathbb{A}$. In the $\varepsilon$-neighborhood NN classification setting, the LOOE for a point $g(x)$ occurs when the number of training points of
opposite class (to that of $x$) that belong to
$\mathbb{B}_g(x,\varepsilon)$ is more than the number of points of
the same class as $x$ that belong to
$\mathbb{B}_g(x,\varepsilon)$. So, the average LOOE for the
$\varepsilon$-neighborhood NN classifier can be given as
\begin{equation}
\text{LOOE}(g,\varepsilon)=\frac{1}{2}+\frac{1}{2n}\sum^n_{i=1}\text{sgn}\left(\sum_{j:\tau_{ij}=-1}\mu_{g(x_j)}\left(\mathbb{B}_g(x_i,\varepsilon)\right)-\sum_{j\ne
i:\tau_{ij}=1}\mu_{g(x_j)}\left(\mathbb{B}_g(x_i,\varepsilon)\right)\right),
\label{Eq:LOO}
\end{equation}
where sgn is the sign function. Minimizing Eq.~(\ref{Eq:LOO}) over
$g\in\mathscr{G}$ and $\varepsilon>0$ is computationally hard
because of its discontinuous and non-differentiable nature.
Instead, based on the observation that LOOE for a point $g(x_i)$
can be minimized by maximizing
$\sum_{j:\tau_{ij}=1}\mu_{g(x_j)}\left(\mathbb{B}_g(x_i,\varepsilon)\right)$
and minimizing
$\sum_{j:\tau_{ij}=-1}\mu_{g(x_j)}\left(\mathbb{B}_g(x_i,\varepsilon)\right)$,
we therefore minimize the proxy to average LOOE by solving
\begin{equation}
\min_{g\in\mathscr{G},\,\varepsilon>0}\frac{1}{n}\sum^n_{i=1}\left(\sum_{j:\tau_{ij}=1}\mu_{g(x_j)}\left(\mathbb{R}^d\setminus\mathbb{B}_g(x_i,\varepsilon)\right)+\sum_{j:\tau_{ij}=-1}\mu_{g(x_j)}\left(\mathbb{B}_g(x_i,\varepsilon)\right)\right).
\label{Eq:LOOEalone}
\end{equation}
In addition, to avoid over-fitting to the training set, the
complexity of $\mathscr{G}$ has to be controlled, for which a
penalty functional, $\Omega[g]$, where
$\Omega:\mathscr{G}\rightarrow\mathbb{R}$ is introduced in
Eq.~(\ref{Eq:LOOEalone}) resulting in the minimization of the
following regularized error functional,
\begin{equation}
\min_{g\in\mathscr{G},\,\varepsilon>0}\,\,\frac{1}{n}\sum^n_{i=1}\left(\sum_{j:\tau_{ij}=1}\mu_{g(x_j)}\left(\mathbb{R}^d\setminus\mathbb{B}_g(x_i,\varepsilon)\right)+\sum_{j:\tau_{ij}=-1}\mu_{g(x_j)}\left(\mathbb{B}_g(x_i,\varepsilon)\right)\right)+\lambda\,\Omega[g],
\label{Eq:LOO-reg}
\end{equation}
with $\lambda>0$ being the regularization parameter. For a given
$x_i$, the above functional minimizes (i) the number of points
with $\tau_{ij}=1$ that do not belong to
$\mathbb{B}_g(x_i,\varepsilon)$, (ii) the number of points with
$\tau_{ij}=-1$ that belong to $\mathbb{B}_g(x_i,\varepsilon)$ and
(iii) the penalty functional. With a few algebraic manipulations,
Eq.~(\ref{Eq:LOO-reg}) can be reduced to
\begin{equation}
\min_{g\in\mathscr{G},\,\varepsilon>0}\,\,\sum^n_{i,j=1}\mu_{g(x_j)}\left(\{g(x):\tau_{ij}\rho^2_g(x_i,x)-\tau_{ij}\varepsilon\ge
0\}\right)+\tilde{\lambda}\,\Omega[g]. \label{Eq:LOO-reg-final}
\end{equation}
where $\tilde{\lambda}=n\lambda$. The first term in
Eq.~(\ref{Eq:LOO-reg-final}) represents the $0$--$1$ loss
function, which is hard to minimize because of its discontinuity
at $0$. Usually, the $0$--$1$ loss is replaced by other loss
functions like hinge/square/logistic loss which are convex. Since
hinge loss is the tightest convex upper bound to the $0$--$1$
loss, we use it as an approximation to the $0$--$1$ loss resulting
in the following minimization problem,
\begin{equation}
\min_{g\in\mathscr{G},\,\varepsilon>0}\,\,\sum^n_{i,j=1}\left[1+\tau_{ij}||g(x_i)-g(x_j)||^2_2-\tau_{ij}\varepsilon\right]_++\tilde{\lambda}\,\Omega[g],
\label{Eq:LOO-reg-final-1}
\end{equation}
where $[a]_+=\max(0,a)$. It is to be noted that
Eq.~(\ref{Eq:LOO-reg-final-1}) is convex in $\varepsilon$ whereas
it is non-convex in $g$ (even when $g$ is linear). So, the
approximation of $0$--$1$ loss with the hinge loss in this case
does not yield a convex program unlike in popular machine learning
algorithms like SVM. Eq.~(\ref{Eq:LOO-reg-final-1}) has an
interesting geometrical interpretation that the points of same
class are closer to one another than to any point from other
classes. This means that the training points are clustered
together according to their class labels, which will definitely
improve the accuracy of NN classifier. However, in comparison to
the method in~\cite{Kilian-05}, such a behavior might be
computationally difficult to achieve. The idea in~\cite{Kilian-05}
is to keep the target neighbors\footnote{The target neighbors are
known \emph{a priori} and are determined by assuming the Euclidean
distance metric.} closer to one another and separate them by a
large margin from the neighbors with non-matching labels. This
method, therefore, does not look beyond target neighbors and
optimizes the Mahalanobis distance metric locally leading to a
global metric. But, the advantage with our formulation is that no
side information (regarding target neighbors) is needed unlike
in~\cite{Kilian-05}. If the underlying metric in $\mathscr{X}$ is
not known, target neighbors cannot be computed and so we do away
with the target neighbor formulation and study the clustering
formulation. Another reason the clustering formulation is
interesting is that it neatly yields a setting to prove a
representer theorem~\cite{Kimeldorf-71} for the
$\varepsilon$-neighborhood NN classification when
$\mathscr{G}=\mathscr{H}_k$ and $\Omega[g]=||g||^2_\mathscr{G}$,
which is discussed in \S\ref{Sec:RKHS}.\par Solving
Eq.~(\ref{Eq:LOO-reg-final-1}) is not easy unless some assumptions
about $\mathscr{G}$ are made. In \S\ref{Sec:RKHS}, we assume
$\mathscr{G}$ as a RKHS with the reproducing kernel $k$ and solve
for a function that optimizes Eq.~(\ref{Eq:LOO-reg-final-1}). In
\S\ref{Sec:Mercer}, we restrict $\mathscr{G}$ to the set of Mercer
kernel maps and show the equivalence between learning the Mercer
kernel map and learning the Mercer kernel.
\section{Regularization in reproducing kernel Hilbert space}\label{Sec:RKHS}
Many machine learning algorithms like SVMs, regularization
networks and logistic regression can be derived within the
framework of regularization in RKHS by choosing the appropriate
empirical risk functional with the penalizer being the squared
RKHS norm~\cite{Evgeniou-00}. In Eq.~(\ref{Eq:LOO-reg-final-1}),
we have extended the regularization framework to
$\varepsilon$-neighborhood NN classification, wherein
$g\in\mathscr{G}$ and $\varepsilon>0$ that minimize the surrogate
to average LOOE have to be computed. Instead of considering any
arbitrary $\mathscr{G}$, we introduce a special structure to
$\mathscr{G}$ by assuming it to be a RKHS, $\mathscr{H}_k$ with
the reproducing kernel $k$. From now onwards, we change the
notation from $\mathscr{G}$ to $\mathscr{H}_k$ and search for
$f\in\mathscr{H}_k$. For the time being, let us assume that
$\mathscr{H}_k=\{f\,|\, f:\mathscr{X}\rightarrow\mathbb{R}\}$. The
penalty functional in Eq.~(\ref{Eq:LOO-reg-final-1}) for
$\mathscr{H}_k$ is defined to be the squared RKHS norm, i.e.,
$\Omega[f]=||f||^2_{\mathscr{H}_k}$.
Eq.~(\ref{Eq:LOO-reg-final-1}) can therefore be rewritten as
\begin{equation}
\min_{f\in\mathscr{H}_k,\,\varepsilon>0}\,\,\sum^n_{i,j=1}\left[1+\tau_{ij}||f(x_i)-f(x_j)||^2_2-\tau_{ij}\varepsilon\right]_++\tilde{\lambda}\,||f||^2_{\mathscr{H}_k}.
\label{Eq:RKHS-f}
\end{equation}
The following lemma provides a representation for $f$ that
minimizes Eq.~(\ref{Eq:RKHS-f}). Using this result,
Theorem~\ref{Thm:Multiout} provides a representation for
$f:\mathscr{X}\rightarrow\mathbb{R}^d$ that minimizes
Eq.~(\ref{Eq:RKHS-f}).
\begin{lem}
\label{Lem:Representer} If $f$ is an optimal solution to
Eq.~(\ref{Eq:RKHS-f}), then $f$ can be expressed as
$f=\sum^n_{i=1}c_ik(.,x_i)$ with $\{c_i\}^n_{i=1}\in\mathbb{R}$
and $\sum^n_{i=1}c_i=0$.
\end{lem}\vspace{-2.5mm}
\begin{proof}
Since $f\in\mathscr{H}_k$, $f(x)=\langle
f,k(.,x)\rangle_{\mathscr{H}_k}$. Therefore, Eq.~(\ref{Eq:RKHS-f})
can be written as
\begin{equation}
\min_{f\in\mathscr{H}_k,\,\varepsilon>0}\,\,\sum^n_{i,j=1}\left[1+\tau_{ij}(\langle
f,k(.,x_i)-k(.,x_j)\rangle_{\mathscr{H}_k})^2-\tau_{ij}\varepsilon\right]_++\tilde{\lambda}\,\langle
f,f\rangle_{\mathscr{H}_k}. \label{Eq:RKHS-thm-1}
\end{equation}
We may decompose $f\in\mathscr{H}_k$ into a part contained in the
span of the kernel functions $\{k(.,x_i)-k(.,x_j)\}^n_{i,j=1}$,
and the one in the orthogonal complement;
$f=f_{||}+f_{\perp}=\sum^{n}_{i,j=1}\alpha_{ij}(k(.,x_i)-k(.,x_j))+f_{\perp}$.
Here $\alpha_{ij}\in\mathbb{R}$ and $f_{\perp}\in\mathscr{H}_k$
with $\langle
f_{\perp},k(.,x_i)-k(.,x_j)\rangle_{\mathscr{H}_k}=0$ for all
$i,j\in\{1,2,\ldots,n\}$. Therefore, $f(x_i)-f(x_j)=\langle
f,k(.,x_i)-k(.,x_j)\rangle_{\mathscr{H}_k}=\langle
f_{||},k(.,x_i)-k(.,x_j)\rangle_{\mathscr{H}_k}=\sum^{n}_{p,m=1}\alpha_{pm}(k(x_i,x_p)-k(x_j,x_p)-k(x_i,x_m)+k(x_j,x_m))$.
Now, consider the penalty functional, $\langle
f,f\rangle_{\mathscr{H}_k}$. For all $f_{\perp}$, $\langle
f,f\rangle_{\mathscr{H}_k}=||f_{||}||^2_{\mathscr{H}_k}+||f_{\perp}||^2_{\mathscr{H}_k}\ge
||\sum^n_{i,j=1}\alpha_{ij}(k(.,x_i)-k(.,x_j))||^2_{\mathscr{H}_k}$.
Thus for any fixed $\alpha_{ij}\in\mathbb{R}$,
Eq.~(\ref{Eq:RKHS-thm-1}) is minimized for $f_{\perp}=0$.
Therefore, the minimizer of Eq.~(\ref{Eq:RKHS-f}) has the form
$f=\sum^n_{i,j=1}\alpha_{ij}(k(.,x_i)-k(.,x_j))$, which is
parameterized by $n^2$ parameters of $\{\alpha_{ij}\}^n_{i,j=1}$.
However, $f$ can be represented by $n$ parameters as
$f=\sum^n_{i=1}c_ik(.,x_i)$ where $\mathbb{R}\ni
c_i=\sum^n_{j=1}(\alpha_{ij}-\alpha_{ji})$ and
$\sum^n_{i=1}c_i=0$.
\end{proof}
\begin{thm}[Multi-output regularization]\label{Thm:Multiout} Let
$\mathscr{H}_k=\{f\,|\,f:\mathscr{X}\rightarrow\mathbb{R}^d\}$. If
$f$ is an optimal solution to Eq.~(\ref{Eq:RKHS-f}), then
\begin{equation}f=\sum^n_{i=1}\mathbf{c}_ik(.,x_i)\label{Eq:Multiout}\end{equation} with
$\mathbf{c}_i\in\mathbb{R}^d$, $\forall i\in\{1,2,\ldots,n\}$ and
$\sum^n_{i=1}\mathbf{c}_i=\mathbf{0}$.
\end{thm}\vspace{-2.5mm}
\begin{proof}
Let
$\tilde{\mathscr{H}}_{\tilde{k}}=\{\tilde{f}\,|\,\tilde{f}:\mathscr{X}\rightarrow\mathbb{R}\}$
with $\tilde{k}$ as its reproducing kernel. Construct
$\mathscr{H}_k=\tilde{\mathscr{H}}_{\tilde{k}}\times
\tilde{\mathscr{H}}_{\tilde{k}}\times\stackrel{d}\ldots\times
\tilde{\mathscr{H}}_{\tilde{k}}=\{(\tilde{f}_1,\tilde{f}_2,\ldots,\tilde{f}_d)\,|\,\tilde{f}_1\in\tilde{\mathscr{H}}_{\tilde{k}},
\tilde{f}_2\in\tilde{\mathscr{H}}_{\tilde{k}},\ldots,\tilde{f}_d\in\tilde{\mathscr{H}}_{\tilde{k}}\}$.
Now, $\mathscr{H}_k$ is a RKHS with the reproducing kernel,
$k=(\tilde{k},\tilde{k},\stackrel{d}\ldots,\tilde{k})$. Then, with
$||f(x)||^2_2=\sum^d_{m=1}||\tilde{f}_m(x)||^2_2=\sum^d_{m=1}(\langle
\tilde{f}_m,\tilde{k}(.,x)\rangle_{\tilde{\mathscr{H}}_{\tilde{k}}})^2$
and $\langle f,f\rangle_{\mathscr{H}_k}=\sum^d_{m=1}\langle
\tilde{f}_m,\tilde{f}_m\rangle_{\tilde{\mathscr{H}}_{\tilde{k}}}$,
Eq.~(\ref{Eq:RKHS-f}) reduces to
\begin{equation}
\min_{\{\tilde{f}_m\}^d_{m=1}\in\tilde{\mathscr{H}}_{\tilde{k}},\,\varepsilon>0}\,\,\sum^n_{i,j=1}\left[1+\tau_{ij}\sum^d_{m=1}(\langle
\tilde{f}_m,\tilde{k}(.,x_i)-\tilde{k}(.,x_j)\rangle_{\tilde{\mathscr{H}}_{\tilde{k}}})^2-\tau_{ij}\varepsilon\right]_++\tilde{\lambda}\,\sum^d_{m=1}\langle
\tilde{f}_m,\tilde{f}_m\rangle_{\tilde{\mathscr{H}}_{\tilde{k}}}.\nonumber
\label{Eq:RKHS-f-multi}
\end{equation}
Applying Lemma~\ref{Lem:Representer} independently to each
$\tilde{f}_m,\,m=1,2,\ldots,d$ proves the
result.\footnote{See~\cite[\S 4.7]{Scholkopf-02} for more
details.}
\end{proof}
We now study the above result for linear kernels. The following
corollary shows that applying a linear kernel is equivalent to
assuming the underlying distance metric in $\mathscr{X}$ to be the
Mahalanobis distance.
\begin{cor}[Linear kernel]
Let $\mathscr{X}=\mathbb{R}^D$ and
$\mathbf{x},\mathbf{y}\in\mathscr{X}$. If
$k(\mathbf{x},\mathbf{y})=\langle
\mathbf{x},\mathbf{y}\rangle=\mathbf{x}^T\mathbf{y}$, then
$\mathbb{R}^d\ni f(\mathbf{x})=\mathbf{Lx}$ and
$\rho_f(\mathbf{x},\mathbf{y})=\sqrt{(\mathbf{x}-\mathbf{y})^T\mathbf{M}(\mathbf{x}-\mathbf{y})}$
with $\mathbf{M}=\mathbf{L}^T\mathbf{L}.$ \label{Corollary: Linear
kernel}
\end{cor}\vspace{-2.5mm}
\begin{proof}
From Eq.~(\ref{Eq:Multiout}), we have
$f(\mathbf{x})=\sum^n_{i=1}\mathbf{c}_i\langle
\mathbf{x},\mathbf{x}_i\rangle=\mathbf{Lx}$, where
$\mathbf{L}=\sum^n_{i=1}\mathbf{c}_i\mathbf{x}^T_i$. Therefore,
$\rho_f(\mathbf{x},\mathbf{y})=||\mathbf{Lx}-\mathbf{Ly}||_2=\sqrt{(\mathbf{x}-\mathbf{y})^T\mathbf{M}(\mathbf{x}-\mathbf{y})}$
with $\mathbf{M}=\mathbf{L}^T\mathbf{L}$.
\end{proof}\vspace{-1mm}
This means that most of the prior work has explored only the
linear kernel. However, it has to be mentioned that~\cite{Vert-06}
derived a dual problem to Eq.~(\ref{Eq:RKHS-f}) by assuming
$\mathscr{X}=\mathbb{R}^D$, $f(\mathbf{x})=\mathbf{Lx}$ and
$\Omega[f]=||\mathbf{\mathbf{L}^T\mathbf{L}}||^2_F$, which is then
kernelized by using the kernel trick.~\cite{Shalev-04} studied the
problem in the same framework as~\cite{Vert-06} barring the
penalty functional but in an online mode. Though our objective
function in Eq.~(\ref{Eq:RKHS-f}) is similar to the one
in~\cite{Shalev-04,Vert-06}, we solve it in a completely different
setting by appealing to regularization in RKHS and without making
any assumptions about $\mathscr{X}$ or its underlying distance
metric. Recently, a different method is proposed by~\cite{Torresani-06},
which kernelized the optimization problem studied
in~\cite{Kilian-05} by assuming a particular parametric form for
$\mathbf{L}$ to invoke the kernel trick. Our method does not make any such assumptions except to restrict $f$ to $\mathscr{H}_k$. Substituting Eq.~(\ref{Eq:Multiout}) in Eq.~(\ref{Eq:RKHS-f}), we get
\begin{eqnarray}
\label{Eq:fsubstituted} \min_{\mathbf{C},\,\varepsilon>0} && \sum^n_{i,j=1}\left[1+\tau_{ij}\text{tr}(\mathbf{CA}_{ij}\mathbf{C}^T)-\tau_{ij}\varepsilon\right]_++\tilde{\lambda}\,\text{tr}(\mathbf{CKC}^T)\nonumber\\
\text{s.t.} &&
\mathbf{C1}=\mathbf{0},\,\mathbf{C}\in\mathbb{R}^{d\times n}
\end{eqnarray}
where
$\mathbf{C}=[\mathbf{c}_1,\mathbf{c}_2,\ldots,\mathbf{c}_n]$,
$\mathbf{K}$ is the kernel matrix with
$\mathbf{K}_{ij}=k(x_i,x_j)$ and
$\mathbf{A}_{ij}=(\mathbf{k}_i-\mathbf{k}_j)(\mathbf{k}_i-\mathbf{k}_j)^T$
with $\mathbf{k}_i$ being the $i^{th}$ column of $\mathbf{K}$. The
objective in Eq.~(\ref{Eq:fsubstituted}) involves terms that are
quadratic and piece-wise quadratic in $\mathbf{C}$, while
piece-wise linear in $\varepsilon$. The constraints are linear in
$\mathbf{C}$ and $\varepsilon$. So, a cursory look might suggest
Eq.~(\ref{Eq:fsubstituted}) to be a convex program. But, it is
actually non-convex because of the presence of $\tau_{ij}=-1$ for
some $i$ and $j$. To make the objective convex, we linearize the
quadratic functions by rewriting Eq.~(\ref{Eq:fsubstituted}) in
terms of $\bar{\mathbf{C}}=\mathbf{C}^T\mathbf{C}$. But this
results in a hard non-convex constraint,
$\text{rank}(\bar{\mathbf{C}})=d$. We relax the rank constraint to
obtain the convex semidefinite program (SDP),
\begin{eqnarray}\min_{\bar{\mathbf{C}},\,\varepsilon>0} && \sum^n_{i,j=1}\left[1+\tau_{ij}\text{tr}(\mathbf{A}_{ij}\bar{\mathbf{C}})-\tau_{ij}\varepsilon\right]_++\tilde{\lambda}\,\text{tr}(\mathbf{K}\bar{\mathbf{C}})\nonumber\\
\text{s.t.}&&\mathbf{1}^T\bar{\mathbf{C}}\mathbf{1}=0,\,\bar{\mathbf{C}}\succeq
0.\end{eqnarray} By introducing slack variables, this SDP can
be written as\vspace{1mm}\setlength{\fboxsep}{1.5mm}
\begin{equation}\label{Eq:rankconstrained}\fbox{$
\begin{array}{rl}
\min_{\bar{\mathbf{C}},\,\{\xi_{ij}\}^n_{i,j=1},\,\varepsilon} &
\langle\bar{\mathbf{C}},\mathbf{K}\rangle_F+\eta\,\sum^n_{i,j=1}\xi_{ij}\\
\text{s.t.} &
\tau_{ij}\left(\langle\bar{\mathbf{C}},-\mathbf{A}_{ij}\rangle_F+\varepsilon\right)\ge
1-\xi_{ij},\,\forall\,i,j\\
&
\langle\bar{\mathbf{C}},\mathbf{11}^T\rangle_F=0,\,\bar{\mathbf{C}}\succeq
0,\,\varepsilon>0\\
&\xi_{ij}\ge 0,\,\forall\,i,j.
\end{array}$}\vspace{1mm}\end{equation}
where $\langle\mathbf{A},\mathbf{B}\rangle_F=\text{tr}(\mathbf{A}^T\mathbf{B})$
and $\eta=1/\tilde{\lambda}$. An interesting observation is that
solving Eq.~(\ref{Eq:rankconstrained}) is equivalent to computing
the Mahalanobis metric, $\bar{\mathbf{C}}$ in $\mathbb{R}^n$ using
the training set, $\{\mathbf{k}_i,y_i\}^n_{i=1}$. This is because,
$\rho_f(x_i,x_j)=||f(x_i)-f(x_j)||_2=\sqrt{(\mathbf{k}_i-\mathbf{k}_j)^T\bar{\mathbf{C}}(\mathbf{k}_i-\mathbf{k}_j)}=\sqrt{\text{tr}(\bar{\mathbf{C}}\mathbf{A}_{ij})}$.
Now, to classify a test point, $x_t$,  $\bar{\mathbf{C}}$ and
$\varepsilon$ obtained by solving Eq.~(\ref{Eq:rankconstrained})
are used to compute
$||f(x_t)-f(x_i)||^2_2=\text{tr}(\bar{\mathbf{C}}\mathbf{A}_{ti}),\,\forall\,i\in\{1,2,\ldots,n\}$
where
$\mathbf{A}_{ti}=(\mathbf{k}_t-\mathbf{k}_i)(\mathbf{k}_t-\mathbf{k}_i)^T$
with $\mathbf{k}_t=\left[k(x_t,x_1),\ldots,k(x_t,x_n)\right]^T$
and the classification is done by either $k$-NN or
$\varepsilon$-neighborhood~NN. \par A careful observation of
Eq.~(\ref{Eq:rankconstrained}) shows an interesting similarity to
the soft-margin formulation of linear binary SVM classifier
wherein a hyperplane in $\mathbb{R}^{n\times n}$ (or
$\mathbb{R}^{n^2}$ by vectorizing the matrices) that separates the
training data, $\{(-\mathbf{A}_{ij},\tau_{ij})\}^n_{i,j=1}$ has to
be computed. The hyperplane is defined by the normal,
$\bar{\mathbf{C}}\in\mathbb{S}^n_+\cap
\{\mathbf{B}:\langle\mathbf{B},\mathbf{11}^T\rangle=0\}$ and its
offset from the origin, $\varepsilon\in\mathbb{R}_{++}$. The
objective function is a trade-off between maximizing the kernel
mis-alignment (between $\bar{\mathbf{C}}$ and $\mathbf{K}$) and
minimizing the training error. $\eta=\infty$ results in a
hard-margin binary SVM classifier. \par While
Eq.~(\ref{Eq:rankconstrained}) can be solved by general purpose
solvers, they scale poorly in the number of constraints, which in
our case is $O(n^2)$. So, we implemented our own special-purpose
solver based on the one developed in~\cite{Kilian-05}, which
exploits the fact that most of the slack variables,
$\{\xi_{ij}\}^n_{i,j=1}$ never attain positive values resulting in
very few active constraints. The solver follows a very simple
two-step update rule: It first takes a step along the gradient to
minimize the objective and then projects $\bar{\mathbf{C}}$ and
$\varepsilon$ onto the feasible set.
\section{Mercer kernel map}\label{Sec:Mercer} As aforementioned, our objective is to reduce the average LOOE of NN classifier by embedding the data
into a Euclidean metric space. One obvious choice of such a
mapping is the Mercer kernel map,
$\phi:\mathscr{X}\rightarrow\ell^N_2$, that satisfies $\langle
\phi(x),\phi(y)\rangle=k(x,y)$, where $k$ is the Mercer kernel.
So, to solve Eq.~(\ref{Eq:LOO-reg-final-1}), we restrict ourselves
to $\mathscr{P}=\{\phi\,|\,\phi:\mathscr{X}\rightarrow\ell^N_2\}$.
Replacing $\mathscr{G}$ by $\mathscr{P}$ and $g$ by $\phi$ in
Eq.~(\ref{Eq:LOO-reg-final-1}), we have
$||\phi(x_i)-\phi(x_j)||^2_2=k(x_i,x_i)+k(x_j,x_j)-2k(x_i,x_j)$.
Defining $k_{ij}=k(x_i,x_j)$, Eq.~(\ref{Eq:LOO-reg-final-1})
reduces to the kernel learning problem,
\begin{equation}
\min_{\mathbf{K}\succeq
0,\,\varepsilon>0}\,\,\sum^n_{i,j=1}\left[1+\tau_{ij}(k_{ii}+k_{jj}-2k_{ij})-\tau_{ij}\varepsilon\right]_++\tilde{\lambda}\,||\mathbf{K}||^2_F,
\label{Eq:LOO-mercer}
\end{equation}
with the penalty functional chosen to be $||\mathbf{K}||^2_F$. It
is to be noted that the embedding is uniquely determined by the
kernel matrix $\mathbf{K}$ and not by $\phi$ as there may exist
$\phi_1$ and $\phi_2$ such that $\phi_1\ne\phi_2$ everywhere but
$k(x,y)=\langle\phi_i(x),\phi_i(y)\rangle$, for $i=1,2$.
Eq.~(\ref{Eq:LOO-mercer}) is a kernel learning problem which is
convex in $\mathbf{K}$ and $\varepsilon$. It depends only on the
entries of the kernel matrix and $\{\tau_{ij}\}^n_{i,j=1}$ while
not utilizing the training data. For sufficiently small
$\tilde{\lambda}$, It can be shown that $\varepsilon=1$ and
$k_{ii}+k_{jj}-2k_{ij}=1-\tau_{ij}$. Therefore, there exists
$\phi$ that achieves
zero LOOE. However, the obtained mapping or $\mathbf{K}$ is not interesting
as it does not allow for an out-of-sample extension and can be
used only in a transductive setting. To extend this method to an
inductive setting, the kernel matrix $\mathbf{K}$ can be
approximated as a linear combination of kernel matrices whose
kernel functions are known~\cite{Lanckriet-04}. This reduces to
minimizing Eq.~(\ref{Eq:LOO-mercer}) over the coefficients of
kernel matrices rather than $\mathbf{K}$. Let us assume that
$\mathbf{K}=\sum^q_{r=1}\beta_r\mathbf{K}_r$ with
$\{\beta_r\}^q_{r=1}\in\mathbb{R}$. Then,
Eq.~(\ref{Eq:LOO-mercer}) reduces to \begin{eqnarray}
\min_{\{\beta_r\}^q_{r=1}\in\mathbb{R},\,\varepsilon>0} && \sum^n_{i,j=1}\left[1+\tau_{ij}\sum^q_{r=1}\beta_r(k^r_{ii}+k^r_{jj}-2k^r_{ij})-\tau_{ij}\varepsilon\right]_++\tilde{\lambda}\sum^q_{r,s=1}\beta_r\beta_s\text{tr}(\mathbf{K}_r\mathbf{K}_s)\nonumber \\ \text{s.t.} && \sum^q_{r=1}\beta_r\mathbf{K}_r\succeq 0, \label{Eq:Mercer-SDP}\end{eqnarray}
which is a SDP with $k^r_{ij}=[\mathbf{K}_r]_{ij}$. By
constraining $\{\beta_r\}^q_{r=1}\in\mathbb{R}_+$, Eq.~(\ref{Eq:Mercer-SDP}) reduces to a quadratic program (QP) given by
\begin{equation}
\min_{\{\beta_r\}^q_{r=1}\in\mathbb{R}_+,\,\varepsilon>0}\,\,\,\,\,\,\,\,\,\,\sum^n_{i,j=1}\left[1+\tau_{ij}\sum^q_{r=1}\beta_r(k^r_{ii}+k^r_{jj}-2k^r_{ij})-\tau_{ij}\varepsilon\right]_++\tilde{\lambda}\,\sum^q_{r,s=1}\beta_r\beta_s\text{tr}(\mathbf{K}_r\mathbf{K}_s).\label{Eq:Mercer-QP}\end{equation}
Though the QP formulation in Eq.~(\ref{Eq:Mercer-QP}) is computionally cheap compared to the SDP formulation in Eq.~(\ref{Eq:Mercer-SDP}), it is not interesting for kernels of the form $h(||\mathbf{x}-\mathbf{y}||^2_2)$ where
$\mathbf{x},\mathbf{y}\in\mathbb{R}^D$ and $h$ is completely
monotonic.\footnote{See~\cite[\S 2.4]{Scholkopf-02} for details on
conditionally positive definite kernels and completely monotonic
functions.} The reason is that, to classify a test point, $x_t$,
we compute $||\phi(x_t)-\phi(x_i)||^2_2=\sum^q_{r=1}\beta_r(
k^r(x_t,x_t)+k^r(x_i,x_i)-2k^r(x_t,x_i))\propto
-2\sum^q_{r=1}\beta_rk^r(x_t,x_i),\,\forall\,i$. Therefore,
minimizing $||\phi(x_t)-\phi(x_i)||^2_2$ is equivalent to
maximizing $\sum^q_{r=1}\beta_rk^r(x_t,x_i)$ and so the QP
formulation yields the same result as that obtained when NN
classification is performed in $\mathscr{X}$. Therefore, this
result eliminates the popular Gaussian kernel from consideration.
However, it is not clear how the QP formulation behaves for other
kernels, e.g., polynomial kernel of degree $\gamma$ and deserves
further study. \par We derived a SDP formulation in
\S\ref{Sec:RKHS} that embeds the training data into a Euclidean
space while minimizing the LOOE of the $\varepsilon$-neighborhood
NN classifier. Since the SDP formulation derived in this section
is based on the approximation of kernel matrix, we prefer the
formulation in \S\ref{Sec:RKHS} to the one derived here for
experiments in \S\ref{Sec:Results}. The purpose of this section is
to show that the metric learning for NN classification can be
posed as a kernel learning problem, which has not been explored
before. Presently, we feel that this framework provides only a
limited scope to explore because of the issues with out-of-sample
extension. However, the derived SDP and QP formulations merit
further study as they can be used for heterogenous data
integration in NN setting similar to the one in SVM
setting~\cite{Lanckriet-04}.
\section{Experiments \& Results}\label{Sec:Results}
In this section, we illustrate the effectiveness of the proposed
method, which we refer to as MENN (metric embedding for nearest
neighbor) in terms of leave-one-out error and generalization error
on four benchmark datasets from the UCI machine learning
repository.\footnote{ftp://ftp.ics.uci.edu/pub/machine-learning-databases}
Since LMNN\footnote{LMNN software is available at
\texttt{http://www.seas.upenn.edu/\textasciitilde kilianw/Downloads/LMNN.html.}} (large
margin nearest neighbor) algorithm, based on optimizing the
Mahalanobis distance metric and proposed by~\cite{Kilian-05}, has
demonstrated improved performance over the standard NN with
Euclidean distance metric, we include it in our performance
comparison. We also compare our method to standard kernelized NN
classification, i.e., by embedding the data using one of the
standard kernel functions, which we refer to as
\emph{Kernel}-NN.\footnote{\emph{Kernel}-NN is computed as
follows. For each training point, $x_j$, the empirical map w.r.t.
$\{x_i\}^n_{i=1}$ defined as $x_j\mapsto
k(.,x_j)|_{\{x_i\}^n_{i=1}}=(k(x_1,x_i),\ldots,k(x_n,x_i))^T\triangleq\mathbf{k}_i$
is computed. $\{\mathbf{k}_i\}^n_{i=1}$ is considered to be the
training set for the NN classification of empirical maps of the
test data.} This method is also included in the comparison as our
method can be seen as learning a Mahalanobis metric in the
empirical kernel map space. As aforementioned,~\cite{Torresani-06}
proposed a kernelized version of LMNN, referred to as KLMCA
(kernel large margin component analysis), which seems to perform
better than LMNN. However, because of the non-availability of
KLMCA code, we are not able to compare our results with it.
However, comparing the results reported in their paper with the
ones in Table~\ref{tab:results}, we notice that MENN offers
similar improvements (over LMNN) than KLMCA does.
\par The wine, iris, ionosphere and balance data sets from UCI
machine learning repository were considered for experimentation.
Since we are interested in testing the SDP formulation (which scales with the number of data points), we focus our experimentation on problems of not too large size.\footnote{LMNN scales with $D$ for which
preprocessing is done by principal component analysis to reduce
the dimensionality.} For large datasets, local optimization of
Eq.~(\ref{Eq:fsubstituted}) is more computionally attractive than
the convex optimization of Eq.~(\ref{Eq:rankconstrained}). 
However, addressing optimization aspects in more detail is omitted here because of space constraints. The results shown in Table~\ref{tab:results} were averaged over $100$ runs ($10$ runs on Iris and Wine, $5$ runs on Balance and
Ionosphere for MENN because of the complexity of SDP) with
different $70/30$ splits of the data for training and testing.
$15\%$ of the training data was used for validation (required in
LMNN, \emph{Kernel}-NN and MENN). The Gaussian kernel,
$\exp(-\rho||\mathbf{x}-\mathbf{y}||^2)$, was used for
\emph{Kernel}-NN and MENN. The parameters $\rho$ and $\lambda$
were set by cross-validation by searching over
$\rho\in\{2^i\}^4_{-4}$ and $\lambda\in\{10^i\}^3_{-3}$. In all
these methods, $k$-NN classifier with $k=3$ was used for
classification. On all datasets except on wine, for which the
mapping to the high dimensional space seems to hurt performance
(noted similarly by~\cite{Torresani-06}), MENN gives better
classification accuracy than LMNN and the other two methods. The
role of empirical kernel maps is not clear as there is no
consistent behavior between the performance accuracy achieved with
standard NN (\emph{Eucl}-NN in Table~\ref{tab:results}) and
\emph{Kernel}-NN.
\begin{table}[t]
\begin{center}
\caption{$k$-NN classification accuracy on UCI datasets: Balance,
Ionosphere, Iris and Wine. The algorithms compared are standard
$k$-NN with Euclidean distance (\emph{Eucl}-NN),
LMNN~\cite{Kilian-05}, \emph{Kernel}-NN (see the text) and MENN
(proposed method). Mean ($\mu$) and standard deviation ($\sigma$)
of the leave-one-out (LOO) and test (generalization) errors are
reported.}\label{tab:results}\vspace{3mm}
\begin{tabular}{|c|c|cccc|}\hline
Dataset & Algorithm/& \emph{Eucl}-NN & LMNN & \emph{Kernel}-NN & MENN\\
$(n,D,l)$&Error&$\mu\pm\sigma$ &$\mu\pm\sigma$ &$\mu\pm\sigma$ &
$\mu\pm\sigma$
\\\hline
Balance& LOO &$17.81\pm 1.86$ & $11.40\pm 2.89$ & $10.73\pm 1.32$ & $\bm{6.87}\pm 1.69$\\
$(625,4,3)$& Test & $18.18\pm 1.88$ & $11.49\pm 2.57$ & $17.46\pm
2.13$ & $\bm{7.12}\pm 1.93$\\\hline
Ionosphere& LOO & $15.89\pm 1.43$ & $3.50\pm 1.18$ & $2.84\pm 0.80$& $\bm{2.27}\pm 0.76$\\
$(351,34,2)$& Test& $15.95\pm 3.03$ & $12.14\pm 2.92$ & $5.81\pm
2.25$ & $\bm{4.21}\pm 1.96$\\\hline
Iris& LOO & $4.30\pm 1.55$ & $\bm{3.25}\pm 1.15$ & $3.60\pm 1.33$ & $3.33\pm 1.02$\\
$(150,4,3)$& Test & $4.02\pm 2.22$ & $4.11\pm 2.26$ & $4.83\pm
2.47$ & $\bm{3.06}\pm 1.65$
\\\hline
Wine& LOO & $5.89\pm 1.35$ & $\mathbf{0.90}\pm 2.80$ & $4.95\pm 1.35$ & $3.06\pm 1.55$\\
$(178,13,3)$& Test & $6.22\pm 2.70$ & $\mathbf{3.41}\pm 2.10$ &
$7.37\pm 2.82$ & $5.18\pm 2.42$ \\\hline
\end{tabular}
\vspace{-6mm}
\end{center}
\end{table}
\section{Related work}\label{Sec:related work}
We briefly review some relevant work and point out similarities
and differences with our work. The central idea in all the
following reviewed works related to the distance metric learning
for NN classification is that similarly labelled examples should
cluster together and be far away from differently labelled
examples. Three major differences between our work and these works is that
(i) no assumptions are made about the underlying distance metric;
the method can be extended to arbitrary metric spaces, (ii) a
suitable proxy to LOOE is chosen as the objective to be minimized,
which neatly yields a setting to prove the representer-like
theorem for NN classification and (iii) interpretation of metric
learning as a kernel learning problem and as a soft-margin linear
binary SVM classification problem.\par\cite{Xing-02} used SDP to
learn a Mahalanobis distance metric for clustering by minimizing
the sum of squared distances between similarly labelled examples
while lower bounding the sum of distances between differently
labelled examples.~\cite{NCA} proposed neighborhood component
analysis (NCA) which minimizes the probability of error under
stochastic neighborhood assignments using gradient descent to
learn a Mahalanobis distance metric. Inspired
by~\cite{NCA},~\cite{Kilian-05} proposed a SDP algorithm to learn
a Mahalanobis distance metric by minimizing the distance between
predefined target neighbors and separating them by a large margin
from the examples with non-matching labels. Compared to these
methods, our method results in a non-convex program which is then
relaxed to yield a convex SDP. Unlike~\cite{Kilian-05}, our method
does not require prior information about target neighbors.
\par \cite{Shalev-04} proposed an online algorithm for learning a
Mahalanobis distance metric with the cost function very similar to
Eq.~(\ref{Eq:LOO-reg-final-1}) and $g$ being
linear.~\cite{Vert-06} studies exactly in the same setting
as~\cite{Shalev-04} but in a batch mode. The kernelized version in
both these methods involves convex optimization over $n^2$
parameters with a semidefinite constraint on $n\times n$ matrix,
which is similar to our method. To ease the computational
burden,~\cite{Vert-06} neglects the semidefinite constraint and
solves a SVM-type QP.~\cite{Torresani-06} proposed a kernel
version of the algorithm proposed in~\cite{Kilian-05} and assumes
a particular parametric form for the linear mapping to invoke the
kernel trick. Instead of solving a SDP, they use gradient descent
to solve the non-convex program. Similar techniques can be used
for our method, especially for large datasets. But, we prefer the
SDP formulation as we do not know how to choose $d$, whereas the
eigen decomposition of $\bar{\mathbf{C}}$ obtained from SDP would
give an idea about $d$.
\section{Conclusion \& Discussion}
We have proposed two different methods to embed arbitrary metric
spaces into a Euclidean space with the goal to improve the
accuracy of a NN classifier. The first method dealt within the
framework of regularization in RKHS wherein a representer-like
theorem was derived for NN classifiers and parallels were drawn
between the $\varepsilon$-neighborhood NN classifier and the
soft-margin linear binary SVM classifier. Although the primary
focus of this work is to introduce a general theoretical framework
to metric learning for NN classification, we have illustrated our
findings with some benchmark experiments demonstrating that the
SDP algorithm derived from the proposed framework performs better
than a previously proposed Mahalanobis distance metric learning
algorithm. In the second method, by choosing the embedding
function to be a Mercer kernel map, we have shown the equivalence
between Mercer kernel map learning and kernel matrix learning.
Though this method is theoretically interesting, currently it is
not useful for inductive learning as it does not allow for an
out-of-sample extension.\par In the future, we would like to apply
our RKHS based algorithm to data from structured spaces,
especially focussing on applications in bio-informatics. On the
Mercer kernel map front, we would like to study it in more depth
and derive an embedding function that supports an out-of-sample
extension. We would also like to apply this framework for
heterogenous data integration in NN setting.
\bibliographystyle{plain}
\small{\bibliography{Refer}}
\end{document}